%% file: moot_cg2004.tex
\documentclass{elsart}

\usepackage{natbib}

\usepackage{amsmath}
\usepackage{amssymb}
\usepackage{varioref}
\usepackage{epic}
\usepackage{eepic}
\usepackage{proof}
\usepackage{color}
\usepackage{graphicx}

\input{moot_pnmacros}

\newcommand{\ynnnyynyy}[1]{
\begin{picture}(1.5,1.5)(0.35,0)
\put(0.5,0){\textcolor{#1}{\rectangle{1}{1}}}
\put(0,1){\textcolor{#1}{\rectangle{0.5}{0.5}}}
\end{picture}}

\newcommand{\nynynyyny}[1]{
\begin{picture}(1.5,1.5)(0.35,0)
\put(0,0){\textcolor{#1}{\rectangle{0.5}{1}}}
\put(1,0){\textcolor{#1}{\rectangle{0.5}{1}}}
\put(0.5,1){\textcolor{#1}{\rectangle{0.5}{0.5}}}
\end{picture}}
\newcommand{\nnyyynyyn}[1]{
\begin{picture}(1.5,1.5)(0.35,0)
\put(0,0){\textcolor{#1}{\rectangle{1}{1}}}
\put(1,1){\textcolor{#1}{\rectangle{0.5}{0.5}}}
\end{picture}}
\newcommand{\nyyynnnyy}[1]{
\begin{picture}(1.5,1.5)(0.35,0)
\put(0.5,0){\textcolor{#1}{\rectangle{1}{0.5}}}
\put(0.5,1){\textcolor{#1}{\rectangle{1}{0.5}}}
\put(0,0.5){\textcolor{#1}{\rectangle{0.5}{0.5}}}
\end{picture}}
\newcommand{\yynnnyyyn}[1]{
\begin{picture}(1.5,1.5)(0.35,0)
\put(0,0){\textcolor{#1}{\rectangle{1}{0.5}}}
\put(0,1){\textcolor{#1}{\rectangle{1}{0.5}}}
\put(1,0.5){\textcolor{#1}{\rectangle{0.5}{0.5}}}
\end{picture}}
\newcommand{\ynynynyny}[1]{
\begin{picture}(1.5,1.5)(0.35,0)
\put(0,0){\textcolor{#1}{\rectangle{0.5}{0.5}}}
\put(1,0){\textcolor{#1}{\rectangle{0.5}{0.5}}}
\put(0.5,0.5){\textcolor{#1}{\rectangle{0.5}{0.5}}}
\put(0,1){\textcolor{#1}{\rectangle{0.5}{0.5}}}
\put(1,1){\textcolor{#1}{\rectangle{0.5}{0.5}}}
\end{picture}}
\newcommand{\nyynyyynn}[1]{
\begin{picture}(1.5,1.5)(0.35,0)
\put(0.5,0.5){\textcolor{#1}{\rectangle{1}{1}}}
\put(0,0){\textcolor{#1}{\rectangle{0.5}{0.5}}}
\end{picture}}
\newcommand{\ynyynynyn}[1]{
\begin{picture}(1.5,1.5)(0.35,0)
\put(0,0.5){\textcolor{#1}{\rectangle{0.5}{1}}}
\put(1,0.5){\textcolor{#1}{\rectangle{0.5}{1}}}
\put(0.5,0){\textcolor{#1}{\rectangle{0.5}{0.5}}}
\end{picture}}
\newcommand{\yynyynnny}[1]{
\begin{picture}(1.5,1.5)(0.35,0)
\put(0,0.5){\textcolor{#1}{\rectangle{1}{1}}}
\put(1,0){\textcolor{#1}{\rectangle{0.5}{0.5}}}
\end{picture}}
\newcommand{\yyynyyyyy}[1]{
\begin{picture}(1.5,1.5)(0.35,0)
\put(0,1){\textcolor{#1}{\rectangle{1.5}{0.5}}}
\put(0,0){\textcolor{#1}{\rectangle{1.5}{0.5}}}
\put(0.5,0.5){\textcolor{#1}{\rectangle{1}{0.5}}}
\end{picture}}
\newcommand{\yyyynyyyy}[1]{
\begin{picture}(1.5,1.5)(0.35,0)
\put(0,1){\textcolor{#1}{\rectangle{1.5}{0.5}}}
\put(0,0){\textcolor{#1}{\rectangle{1.5}{0.5}}}
\put(0,0.5){\textcolor{#1}{\rectangle{0.5}{0.5}}}
\put(1,0.5){\textcolor{#1}{\rectangle{0.5}{0.5}}}
\end{picture}}
\newcommand{\yyyyynyyy}[1]{
\begin{picture}(1.5,1.5)(0.35,0)
\put(0,1){\textcolor{#1}{\rectangle{1.5}{0.5}}}
\put(0,0){\textcolor{#1}{\rectangle{1.5}{0.5}}}
\put(0,0.5){\textcolor{#1}{\rectangle{1}{0.5}}}
\end{picture}}

\newcommand{\yyyyyyyyy}[1]{
\begin{picture}(1.5,1.5)(0.35,0)
\put(0,0){\textcolor{#1}{\rectangle{1.5}{1.5}}}
\end{picture}}

\definecolor{gray90}{gray}{0.90}
\definecolor{gray80}{gray}{0.80}
\definecolor{gray70}{gray}{0.70}
\definecolor{gray60}{gray}{0.60}
\definecolor{gray50}{gray}{0.80} 
\definecolor{gray40}{gray}{0.40}
\definecolor{gray30}{gray}{0.30}
\definecolor{gray20}{gray}{0.20}
\definecolor{gray10}{gray}{0.10}
\definecolor{gray95}{gray}{0.95}
\definecolor{gray85}{gray}{0.85}
\definecolor{gray75}{gray}{0.75}
\definecolor{gray65}{gray}{0.65}
\definecolor{gray55}{gray}{0.55}
\definecolor{gray45}{gray}{0.45}
\definecolor{gray35}{gray}{0.35}
\definecolor{gray25}{gray}{0.25}
\definecolor{gray15}{gray}{0.15}
\definecolor{gray05}{gray}{0.05}

\newtheorem{theorem}{Theorem}
\newtheorem{definition}[theorem]{Definition}

\newtheorem{proposition}[theorem]{Proposition}
\newtheorem{corollary}[theorem]{Corollary}

\usepackage{pstricks}
\usepackage{pst-node}

\begin{document}

\begin{frontmatter}



\title{Graph Algorithms for Improving Type-Logical Proof Search}


\author{Richard Moot\corauthref{work}}
\corauth[work]{CNRS and INRIA Futurs}
\address{SIGNES, LaBRI\\351, cours de la
  Lib\'{e}ration\\33405 Talence, France\ead{Richard.Moot@labri.fr}\ead[url]{http://www.labri.fr/Perso/~moot}}

\begin{abstract}
Proof nets are a graph theoretical representation of proofs in various
fragments of type-logical grammar. In spite of this basis in graph
theory, there has been relatively little attention to the use of graph
theoretic algorithms for type-logical proof search.

In this paper we will look at several ways in which standard graph
theoretic algorithms can be used to restrict the search space. In
particular, we will provide an $O(n^4)$ algorithm for selecting
an optimal axiom link at any stage in the proof search as well as an
$O(kn^3)$ algorithm for selecting the $k$ best proof candidates.

\end{abstract}

\begin{keyword}
Automated Deduction \sep Floyd-Warshall Algorithm \sep Lambek Calculus \sep
Proof Net \sep Ranked Assignments

\end{keyword}

\end{frontmatter}

\section{Introduction}
\label{intro}

Type-logical grammar \citep{TLG,M95} is an attractive logical view of
grammatical theory. Advantages of this theory over other formalisms
include a simple, transparent theory of ($\lambda$ term) semantics
thanks to the Curry-Howard isomorphism and effective learning
algorithms for inducing grammars from linguistic data \citep{buspenn}.

Proof nets, first introduced for linear logic by \citet{girard}, are a way of presenting type-logical proofs which circumvents 
the redundancies present in, for example, the sequent calculus by 
performing all logical rules `in parallel'. The only non-determinism in 
trying to prove a theorem consists of selecting pairs of axiom links. Each 
possible selection --- if correct --- will result in a different proof.

However, many of these possible selections can never contribute to a 
proof net, while a naive algorithm might try these selections many times. It is the goal of this paper to provide algorithms for 
filtering out these possibilities at an early stage and selecting 
the axiom link which is most restricted, thereby improving the performance
of proof search.

Given that the problem we are trying to solve is known to be NP complete, even in
the non-commutative case, it would be too much to hope for a
polynomial algorithm \citep{kanovich,pentus03np}. However, we will see an algorithm for computing
the best possible continuation of a partial proof net in $O(n^4)$.

A second aim is to develop a polynomial algorithm by means of 
approximation. If we consider only the best $k$ axiom links, then we can 
find these in $O(kn^3)$. When best is defined as `having axiom links 
with the shortest total distance' this algorithm converges with results on 
proof nets and processing \citep{pnproc,incrpro}.

\section{Proof Nets and Essential Nets}

In this section we will look at two ways of presenting proof
nets for multiplicative intuitionistic linear logic (\textbf{MILL})
together with their correctness criteria and some basic properties.

Though the results will be focused on an associative, commutative
system, it is simple to enforce non-commutativity by demanding
the axiom links to be planar \citep{Roorda} or by labeling, either with
pairs of string positions \citep{hollp} or by algebraic terms
\citep{deGroote}. In order to have more flexibility in dealing with
linguistic phenomena, other constraints on the correctness of proof
nets have been proposed \citep{mp}, but given that the associative,
commutative logic is the worst case (in the sense that it allows the
most axiom links) with respect to other fragments of categorial
grammars there are no problems adapting the results of this paper to
other systems. However, we leave the question of whether it is
possible to perform better for more restricted type-logical grammars open.

The choice of presenting the logic with two implications which differ only 
in the order of the premisses of the links is intended to make the
extensions to the non-commutative case more clearly visible.

\subsection{Sequent Calculus}

Table~\ref{sequentl} shows the sequent calculus for the Lambek 
calculus \textbf{L}, first proposed by \citet{lambek}. The commutative 
version, the Lambek-van Benthem calculus \textbf{LP}, is also known as the 
multiplicative fragment of intuitionistic linear logic
\textbf{MILL}. An example sequent derivation is shown in
Figure~\ref{fig:exseq}.

\begin{table}
\begin{center}
\begin{tabular}{cc}
\infer[\bo \textit{Ax}\bc]{A \ra A}{} &
\infer[\bo \textit{Cut}\bc]{\Gamma ,\Delta,\Gamma^{\prime} \ra C}
                 {\Delta \ra A & \Gamma, A, \Gamma^{\prime} \ra C}\\[3mm]
\end{tabular}

\begin{tabular}{cc}
\infer[\bo \textit{L}\bullet \bc]{\Gamma, A\bullet B, \Delta \ra C}
                 {\Gamma, A, B, \Delta \ra C} &
\infer[\bo \textit{R}\bullet \bc]{\Gamma,\Delta \ra A\bullet B}
                 {\Gamma \ra A & \Delta \ra B} \\[3mm]
\infer[\bo \textit{L}\mathbin{/} \bc]{\Gamma ,A\mathbin{/} B,\Delta,\Gammap \ra C}
                 {\Delta \ra B & \Gamma, A, \Gammap \ra C} &
\infer[\bo \textit{R}\mathbin{/} \bc]{\Gamma \ra A\mathbin{/}B}
                 {\Gamma , B \ra A} \\[3mm]
\infer[\bo \textit{L}\mathbin{\bs} \bc]{\Gamma, \Delta, B\mathbin{\bs} A, \Gammap \ra C}
                   {\Delta \ra B & \Gamma, A, \Gammap \ra C} & 
\infer[\bo \textit{R}\mathbin{\bs} \bc]{\Gamma \ra B\mathbin{\bs} A}
                   {B, \Gamma \ra A} \\[3mm]
\end{tabular}
\end{center}
\caption{The sequent calculus for \textbf{L}/\textbf{MILL} with commutativity implicit}
\label{sequentl}
\end{table}

\begin{figure}
\begin{center}
\mbox{
 \infer[\bo \textit{L}\mathbin{/} \bc]{ s\mathbin{/} (np \mathbin{\bs} s), 
(s \mathbin{/}  (np \mathbin{\bs} s))\mathbin{\bs} s \vdash s}{
     \infer[\bo \textit{R}\bs \bc]{(s \mathbin{/}  (np \mathbin{\bs} 
s))\mathbin{\bs} s \vdash np \mathbin{\bs} s}{
        \infer[\bo \textit{L}\bs \bc]{ np, (s \mathbin{/}  (np 
\mathbin{\bs}s) )\mathbin{\bs} s \vdash s }{
            \infer[\bo \textit{R}/\bc]{ np \vdash s\mathbin{/} (np 
\mathbin{\bs} s)}{
               \infer[\bo \textit{L}\bs\bc]{ np, np \mathbin{\bs} s \vdash s}{
                  \infer[\bo\textit{Ax}\bc]{ np \vdash np}{\rule{0pt}{2ex}}
                & \infer[\bo\textit{Ax}\bc]{ s \vdash s}{}
               }
            }
            & \infer[\bo \textit{Ax}\bc]{s \vdash s}{}
        }
     }
   & \infer[\bo \textit{Ax}\bc]{ s \vdash s}{}
}}
\end{center}
\caption{Example sequent derivation}
\label{fig:exseq}
\end{figure}

\subsection{Proof Nets}

Proof nets are an economic way of presenting proofs for linear logic,
which is particularly elegant for the multiplicative fragment. When 
looking at sequent proofs, there are often many ways of deriving 
essentially `the same' proof. Proof nets, on the other hand, are
inherently redundancy-free.

We define proof nets as a subset of \emph{proof structures}. A proof 
structure is a collection of the links shown in 
Table~\ref{tab:linkle} which satisfies the conditions of Definition~\ref{def:ps}. A link has its conclusions drawn at the bottom and 
its premisses at the top. The axiom link, top left of the table, has no 
premisses and two conclusions which can appear in any order. The cut link, 
top right of the table, is mentioned only for completeness; we will not 
consider cut links in this paper. A cut link has two premisses, which can 
appear in any order, and no conclusions. All other links have an explicit 
left premiss and right premiss. We also distinguish between negative 
(antecedent) and positive (succedent) formulas and between tensor (solid) 
and par (dotted) links.

\begin{table}
\begin{center}
\scalebox{0.8}{
\vspace{2\baselineskip}
\begin{tabular}{c@{\qquad\qquad}c}
\begin{picture}(4,2)
 \put(0,0){\makebox(0,0){$\ant{A}$}}
 \put(4,0){\makebox(0,0){$\suc{A}$}}
 \drawline(0,1)(0,2)(4,2)(4,1)
\end{picture} &
\begin{picture}(4,2)
 \put(0,2){\makebox(0,0){$\ant{A}$}}
 \put(4,2){\makebox(0,0){$\suc{A}$}}
 \drawline(0,1)(0,0)(4,0)(4,1)
\end{picture} \\[10mm]
\end{tabular}}

\scalebox{0.8}{\begin{tabular}{c@{\qquad\qquad}c@{\qquad\qquad}c}
\parlink{\ant{A \bullet B}}{\ant{B}}{\ant{A}} &
\timeslink{\ant{A \mathbin{/} B}}{\suc{B}}{\ant{A}} &
\timeslink{\ant{B \mathbin{\backslash} A}}{\ant{A}}{\suc{B}} \\[10mm]
\timeslink{\suc{A \bullet B}}{\suc{A}}{\suc{B}} &
\parlink{\suc{A \mathbin{/} B}}{\suc{A}}{\ant{B}} &
\parlink{\suc{B \mathbin{\bs} A}}{\ant{B}}{\suc{A}} \\[10mm]
\end{tabular}}

\end{center}
\caption{Links for proof structures}
\label{tab:linkle}
\end{table}

\begin{definition}\label{def:ps} A proof structure $\mathcal S$ is a collection of links 
such that:
\begin{enumerate}
\item every formula is the conclusion of exactly one link,
\item every formula is the premiss of at most one link, formulas which are 
not the premiss of a link are called the \emph{conclusions} of the proof 
structure,
\item a proof structure has exactly one positive conclusion.
\end{enumerate}
\end{definition}

Given a proof structure, we want to decide if it is a proof net, that is, 
if it corresponds to a sequent proof. A correctness criterion allows us to 
accept the proof structures which are correct and reject those which are 
not. In this section, we will look at the acyclicity and connectedness 
condition from \citet{multiplicatives}, which is perhaps the most 
well-known correctness condition for proof nets in multiplicative linear 
logic. We will look at another condition in the next section.

\begin{definition}\label{def:switch}For a proof structure $\cals$, a
{\em switching} $\omega$ for $\cals$ is a choice for
every par link of one of its premisses.
\end{definition}

\begin{definition}\label{def:corgraph}
From a proof structure $\cals$ and a switching $\omega$ we obtain a
{\em correction graph} $\omega\cals$ by replacing
all par links

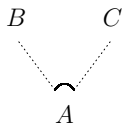
\[
  \scalebox{0.8}{\parlink{A}{B}{C}}
\]

\noindent depending on whether $\omega$ selects the left or the right
premiss of the link,  by one of the following links.

\[ 
   \scalebox{0.8}{\leftlink{A}{B}{C} \hspace{2cm} \rightlink{A}{B}{C}}
\]
\end{definition}

\begin{theorem}[\citet{multiplicatives}]\label{acc} A sequent $\Gamma
  \vdash C$ is provable in \textbf{MILL} iff all correction graphs of
  the corresponding proof structure are acyclic and connected, ie.\ it
  is a \emph{proof net}.
\end{theorem}

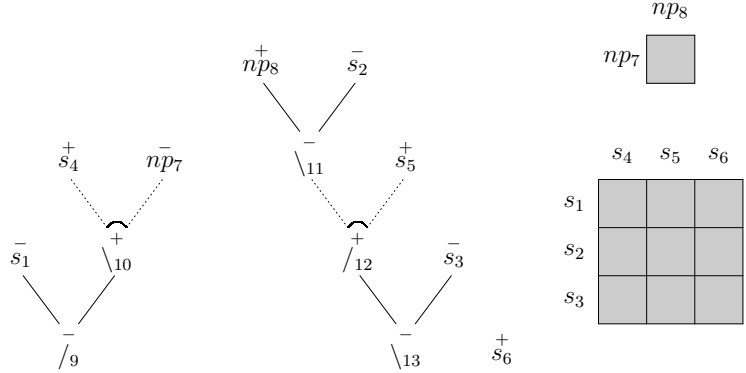
\begin{figure}
 \begin{center}
\scalebox{0.8}{
  \begin{picture}(30,18)
   \put(2,1){\timeslink{\ant{\mathbin{/_9}}}{\ant{s_1}}{\suc{\mathbin{\bs_{10}}}}}
   \put(4,5){\parlink{}{\suc{s_4}}{\ant{np_7}}}

   \put(12,9){\timeslink{\ant{\mathbin{\bs_{11}}}}{\suc{np_8}}{\ant{s_2}}}
   \put(14,5){\parlink{\suc{\mathbin{/_{12}}}}{}{\suc{s_5}}}
   \put(16,1){\timeslink{\ant{\mathbin{\bs_{13}}}}{}{\ant{s_3}}}

   \put(22,1){\makebox(0,0){\suc{s_6}}}

\put(26,2){\textcolor{gray50}{\rectangle{6}{6}}}
\path(26,2)(32,2)
\path(26,4)(32,4)
\path(26,6)(32,6)
\path(26,8)(32,8)
\path(26,2)(26,8)
\path(28,2)(28,8)
\path(30,2)(30,8)
\path(32,2)(32,8)
\put(25,7){\makebox(0,0){$s_1$}}
\put(25,5){\makebox(0,0){$s_2$}}
\put(25,3){\makebox(0,0){$s_3$}}
\put(27,9){\makebox(0,0){$s_4$}}
\put(29,9){\makebox(0,0){$s_5$}}
\put(31,9){\makebox(0,0){$s_6$}}

\put(28,12){\textcolor{gray50}{\rectangle{2}{2}}}
\put(29,15){\makebox(0,0){$np_8$}}
\put(27,13){\makebox(0,0){$np_7$}}
\path(28,12)(28,14)(30,14)(30,12)(28,12)
  \end{picture}}
 \end{center}

\caption{Example proof frame}
\label{fig:pfsomeoneleft}
\end{figure}

Proof search in a proof net system is a rather direct reflection of
the definitions. Given a sequent $\Gamma \vdash C$ we unfold the
negative formulas in $\Gamma$ and the positive formula $C$, giving us
a proof frame. Note that given a polarized formula, exactly
one link will apply, making this stage trivial. An example proof frame for the sequent

\[
(np \mathbin{\bs} s), 
(s \mathbin{/}  (np \mathbin{\bs} s))\mathbin{\bs} s \vdash s
\]

\noindent of Figure~\ref{fig:exseq} is given
in Figure~\vref{fig:pfsomeoneleft}. We have given the atomic formulas an index
as subscript only to make it easier to refer to them; the numbers are
not formally part of the logic.
The matrix next to the proof
frame in the figure represents the possible linkings: the
rows are the negative formula occurrences, whereas the columns are the
positive formula occurrences. White entries represent currently impossible
connections whereas colored entries represent the current possibilities.

\begin{figure}
 \begin{center}
\scalebox{0.8}{
  \begin{picture}(34,18)
   \put(2,1){\timeslink{\ant{\mathbin{/_9}}}{\ant{s_1}}{\suc{\mathbin{\bs_{10}}}}}
   \put(4,5){\parlink{}{\suc{s_4}}{\ant{np_7}}}

   \put(12,9){\timeslink{\ant{\mathbin{\bs_{11}}}}{\suc{np_8}}{\ant{s_2}}}
   \put(14,5){\parlink{\suc{\mathbin{/_{12}}}}{}{\suc{s_5}}}
   \put(16,1){\timeslink{\ant{\mathbin{\bs_{13}}}}{}{\ant{s_3}}}

   \put(22,1){\makebox(0,0){\suc{s_6}}}
   \path(8,10)(8,15)(12,15)(12,14)
   \path(4,10)(4,16)(16,16)(16,14)
   \path(2,6)(2,17)(18,17)(18,10)
   \path(20,6)(20,7)(22,7)(22,2)

\put(28,12){\textcolor{gray50}{\rectangle{2}{2}}}
\put(30,2){\textcolor{gray50}{\rectangle{2}{2}}}
\put(26,4){\textcolor{gray50}{\rectangle{2}{2}}}
\put(28,6){\textcolor{gray50}{\rectangle{2}{2}}}
\path(26,2)(32,2)
\path(26,4)(32,4)
\path(26,6)(32,6)
\path(26,8)(32,8)
\path(26,2)(26,8)
\path(28,2)(28,8)
\path(30,2)(30,8)
\path(32,2)(32,8)
\put(25,7){\makebox(0,0){$s_1$}}
\put(25,5){\makebox(0,0){$s_2$}}
\put(25,3){\makebox(0,0){$s_3$}}
\put(27,9){\makebox(0,0){$s_4$}}
\put(29,9){\makebox(0,0){$s_5$}}
\put(31,9){\makebox(0,0){$s_6$}}

\put(29,15){\makebox(0,0){$np_8$}}
\put(27,13){\makebox(0,0){$np_7$}}
\path(28,12)(28,14)(30,14)(30,12)(28,12)
  \end{picture}}
 \end{center}

\caption{Example proof net}
\label{fig:pnsomeoneleft}
\end{figure}
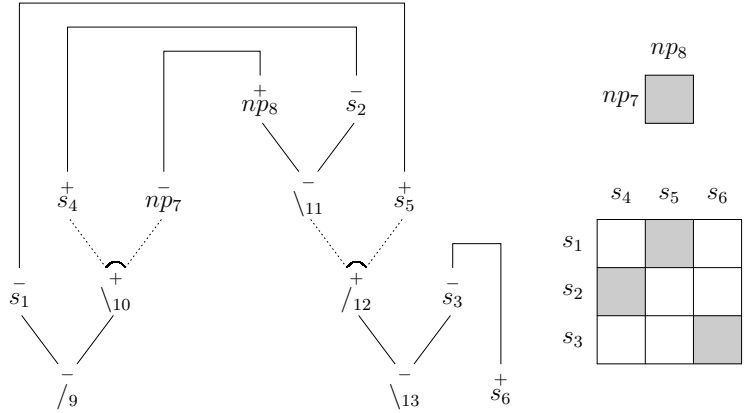

The next stage consists of transforming the proof frame into a proof
structure by linking atomic formulas of opposite polarity. It is this
stage which will concern us in this paper. This is a matter of putting
exactly one mark in every row and every column of the
table. Figure~\ref{fig:pnsomeoneleft} shows one of the 6 possible
linkings next to the matrix which corresponds to it.

Finally, we need to check the correctness condition. Though there are many 
correction graphs for a proof structure, \citet{pnlinear} shows we can 
verify the correctness of a proof structure in linear time. The proof
structure in Figure~\ref{fig:pnsomeoneleft} is indeed a proof net,
which we can verify by testing all correction graphs for acyclicity
and connectedness. Of the 5 alternative linkings, only one other
produces a proof net.

\subsection{Essential Nets}

For out current purposes, we are interested in an alternative
correctness criterion proposed by \citet{essnet}. This criterion is
based on a different way of decomposing a sequent, this time into a
directed graph, with conditions on the paths performing the role of a
correctness criterion. A net like this is called an {\em essential net}. 
The links for essential nets are shown in Table~\vref{tab:dyngraph}, 
though we follow \citet{deGroote} in reversing the arrows of 
\citet{essnet}.

\begin{table}
\begin{center}

\scalebox{0.8}{
\vspace{2\baselineskip}
\begin{tabular}{c@{\qquad\qquad}c}
\begin{picture}(4,2)
 \put(0,0){\makebox(0,0){$\ant{A}$}}
 \put(4,0){\makebox(0,0){$\suc{A}$}}
 \drawline(0,1)(0,2)(4,2)
 \put(4,2){\vector(0,-1){1}}
\end{picture} &

\begin{picture}(4,2)
 \put(0,2){\makebox(0,0){$\ant{A}$}}
 \put(4,2){\makebox(0,0){$\suc{A}$}}
 \drawline(0,0)(4,0)(4,1)
 \put(0,0){\vector(0,1){1}}
\end{picture} \\[10mm]
\end{tabular}}

\scalebox{0.8}{
\begin{tabular}{c@{\qquad\qquad}c@{\qquad\qquad}c}
\trlprod{\ant{A\bullet B}}{\ant{A}}{\ant{B}} &

\trldra{\ant{A \mathbin{/} B}}{\ant{A}}{\suc{B}} &

\trldla{\ant{B \mathbin{\backslash} A}}{\suc{B}}{\ant{A}} \\[10mm]

\trrprod{\suc{A\bullet B}}{\suc{A}}{\suc{B}} &

\trrdl{\suc{A \mathbin{/} B}}{\ant{B}}{\suc{A}} &

\trrdr{\suc{B \mathbin{\bs} A}}{\suc{A}}{\ant{B}} \\[10mm]
\end{tabular}}

\end{center}
\caption{Links for essential nets}
\label{tab:dyngraph}
\end{table}

\begin{definition} Given an essential net $\mathcal E$ its positive
  conclusion is called the {\em output} of the essential net and the
  negative conclusions, as well as the negatives premisses of any
  positive $/$ or $\bs$ link, are called its {\em inputs}.
\end{definition}

\begin{definition}\label{def:correctdyn} An essential net is {\em correct} iff the following
properties hold.

\begin{enumerate}
\item it is acyclic,
\item every path from the negative premiss of a positive $/$ or
$\bs$ link passes through the conclusion of this link,
\item every path from the inputs of the graph passes reaches
the output of the graph.
\end{enumerate}
\end{definition}

\begin{theorem}[\citet{essnet}]\label{corressnet} A sequent $\Gamma \vdash C$ is
provable in \textbf{MILL} iff its essential net is correct.
\end{theorem}

Condition (1) reflects the acyclicity condition on proof nets, whereas
conditions (2) and (3) reflect the connectedness condition. The
formulation of `every path' exists only to ensure correctness of the
negative $\bullet$ link; in all other cases there is at most one
path between two points in a correct essential net.

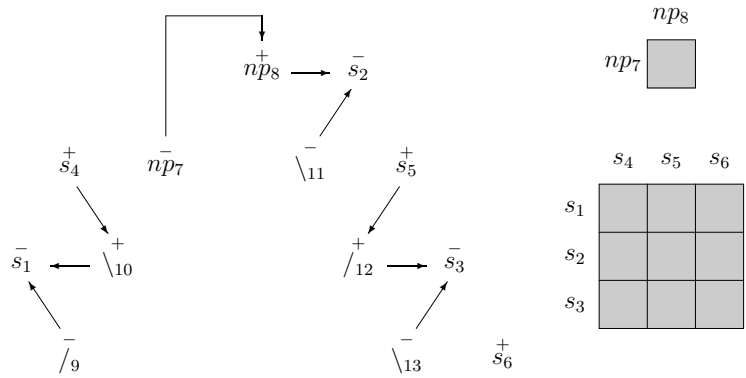
\begin{figure}
 \begin{center}
  \scalebox{0.8}{
  \begin{picture}(34,16)
   \put(2,1){\trldra{\ant{\mathbin{/_9}}}{\ant{s_1}}{\suc{\mathbin{\bs_{10}}}}}
   \put(4,5){\trrdr{}{\suc{s_4}}{\ant{np_7}}}

   \put(12,9){\trldla{\ant{\mathbin{\bs_{11}}}}{\suc{np_8}}{\ant{s_2}}}
   \put(14,5){\trrdl{\suc{\mathbin{/_{12}}}}{}{\suc{s_5}}}
   \put(16,1){\trldla{\ant{\mathbin{\bs_{13}}}}{}{\ant{s_3}}}

   \put(22,1){\makebox(0,0){\suc{s_6}}}
   \path(8,10)(8,15)(12,15)
   \put(12,15){\vector(0,-1){1}}

\put(26,2){\textcolor{gray50}{\rectangle{6}{6}}}
\path(26,2)(32,2)
\path(26,4)(32,4)
\path(26,6)(32,6)
\path(26,8)(32,8)
\path(26,2)(26,8)
\path(28,2)(28,8)
\path(30,2)(30,8)
\path(32,2)(32,8)
\put(25,7){\makebox(0,0){$s_1$}}
\put(25,5){\makebox(0,0){$s_2$}}
\put(25,3){\makebox(0,0){$s_3$}}
\put(27,9){\makebox(0,0){$s_4$}}
\put(29,9){\makebox(0,0){$s_5$}}
\put(31,9){\makebox(0,0){$s_6$}}

\put(28,12){\textcolor{gray50}{\rectangle{2}{2}}}
\put(29,15){\makebox(0,0){$np_8$}}
\put(27,13){\makebox(0,0){$np_7$}}
\path(28,12)(28,14)(30,14)(30,12)(28,12)
  \end{picture}}
 \end{center}

\caption{Example essential net}
\label{fig:ensomeoneleft}
\end{figure}

Figure~\ref{fig:ensomeoneleft} gives the essential net corresponding to the
proof frame we have seen before, but this time with the $np$ axiom
link already performed. Remark that we have simply unfolded the
formulas as before, just with a different set of links. 

It will be our goal to eliminate as many axiom
links as possible for this example.

Though the correctness criterion was originally
formulated for the multiplicative intuitionistic fragment of linear
logic only, \citet{murong} show --- in addition to giving a linear time
algorithm for testing the correctness of an essential net --- that we can
transform a classical proof net into an essential net in linear
time. So our results in the following sections can be applied to the
classical case as well.

\subsection{Basic Properties}

In order to better analyze the properties of the algorithms we
propose, we will first take a look at some basic properties of proof
nets.

\subsubsection{Axiom Links}

Since we will be concerned with finding an axiom linking for a partial
proof structure $\mathcal P$ which will turn $\mathcal P$ into a proof
net, we first given some bounds on the number of proof structures we
will have to consider. Given that the problem we are trying to solve
is NP complete, it is not surprising these bounds are quite high. 

\begin{proposition}\label{prop:countcheck} Let $\mathcal P$ be a proof
  net and $f$ an atomic formula, then the number of positive
  occurrences of $f$ is equal to the number of negative occurrences of $f$.
\end{proposition}

This proposition follows immediately from the fact that every atomic
formula is the conclusion of an axiom link, where each axiom link has
one positive and one negative occurrence of a formula $f$ as its
conclusion.

\begin{proposition} Every proof frame $\mathcal F$ has $O(a!)$ axiom
  linkings which produce a proof structure, where $a$ is the maximum
  number of positive and negative occurrences of an atomic formula in
  $\mathcal F$.
\end{proposition}

If we have $a$ positive atomic formulas, we have $a$ possibilities for
the first one, since all negative formulas may be selected, followed
by $a-1$ for the second etc.\ giving us $a!$ possibilities.

\begin{proposition} Every proof frame $\mathcal F$ has $O(4^a)$ planar axiom
  linkings which produce a proof structure, where $a$ is the maximum
  number of positive and negative occurrences of an atomic formula in
  $\mathcal F$.
\end{proposition}

This follows from the fact that a planar axiom linking is simply a
binary bracketing of the atomic formulas and the fact that there are
$C_{a-1}$ such bracketings, where $C_k$, the $k$th Catalan number,
approaches $4^k/\sqrt{\pi} k^{3/2}$.

\begin{proposition}\label{prop:posax} For every partial proof structure with $a$ atomic
  formulas which are not the conclusion of any axiom link there are
  $O(a^2)$ possible axiom links.
\end{proposition}

Given that every positive atomic formula can be linked to every
negative atomic formula of the same atomic type this gives us $a^2$
pairs.

\subsubsection{Graph Size}

\begin{proposition} For every proof structure $\mathcal S$ with $h$ negative
  conclusions, 1 positive conclusion, $p$ par links and $t$ tensor links, the following
  equation holds.

\[
  p + h = t + 1 = a
\]
\end{proposition}

Given Proposition~\ref{prop:countcheck}, the number of positive and
negative atomic formulas is both $a$. Suppose we want to construct a
proof structure $\mathcal S$ with $h$ negative conclusions and 1
positive conclusion from these atomic formulas. When we look at the
links in Table~\ref{tab:linkle} we see that all par links reduce the
number of negative conclusions by 1 and all tensor links reduce the
number of positive conclusions by 1.

\begin{proposition}\label{prop:graphsize} Every essential net $\mathcal E$ has $v = h + 1 + 2(t +
  p) = O(a)$ vertices and $2t+p \leq e \leq 2(t + p) + a = O(a)$
  edges.
\end{proposition}

This follows immediately from inspection of the links: all conclusions
of the essential net ($h$ negative and 1 positive) start out as a
single vertex and every link adds two new vertices. For the edges: the
minimum number is obtained when we have no axiom links and all par
links are positive links for $\bs$ or $/$ which introduce one edge, the
maximum number includes $a$ axiom links and par links which are all
negative links for $\bullet$.

\begin{corollary} An essential net is \emph{sparse}, ie.\ the number of
  edges is proportional to the number of vertices, but if we add edges
  for all possible axiom links it will be \emph{dense},
  ie.\ $e$ is proportional to $v^2$, 
\end{corollary}

Immediate from Proposition~\ref{prop:posax} and
Proposition~\ref{prop:graphsize}.

\section{Acyclicity}

We begin by investigating the acyclicity condition, condition (1) from
Definition \ref{def:correctdyn}, which is the easiest to verify.

In order to select the axiomatic formula which is most constrained with
respect to the acyclicity condition we can simply enumerate all $a^2$
possible axiom links and reject those which produce a cycle.

We can easily verify whether a graph contains a cycle in time proportional to the
representation of the graph, $v+e$, using either breadth-first search or
depth-first search \citep[e.g.][Section 23.2 and 23.3
  respectively]{algo}, giving us an $O(a^2(v+e)) = O(a^3)$ algorithm
for verifying all pairs.

However, this means we will visit the vertices and edges of the graph
many times. It is therefore a practical improvement to compute the
transitive closure of the graph in advance, after which we can perform the
acyclicity queries in constant time. 

In this paper we will use the Floyd-Warshall algorithm \citep{algo} for computing
the transitive closure, which computes the transitive closure of a
directed graph in $O(v^3)$ time. Though there are algorithms which perform
asymptotically better for sparse graphs, it is hard to beat this
algorithm in practice even for sparse graphs because of the small
constants involved, while for dense graphs, which we will consider in
the next section when we take all possible axiom links into account,
it is the algorithm of choice \citep{sedgewick01graph}.

\begin{figure}
\begin{center}
\begin{pspicture}(2,3)
\rput(1,0.5){\circlenode{A}{a}}
\rput(1,2.5){\circlenode{B}{b}}
\rput(0.2,1.5){\circlenode{C}{c}}
\nccurve{->}{A}{B}
\ncarc{->}{A}{C}
\ncarc{->}{C}{B}
\end{pspicture}
\end{center}
\caption{Eliminating node $c$ from the path from $a$ to $b$}
\label{fig:fwillus}
\end{figure}
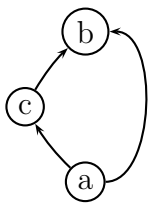

The Floyd-Warshall algorithm is based on successively eliminating the
intermediate vertices $c$ from every path from $a$ to $b$. Given a
vertex $c$ and the paths $a\rightarrow c \rightarrow b$ for all $a$
and $b$ we create a direct path $a\rightarrow b$ if it didn't exist
before. That is to say, there is a path from $a$ to $b$ if either
there is a path from $a$ to $c$ and from $c$ to $b$ or if there is a
path from $a$ to $b$ which we already knew about
(Figure~\vref{fig:fwillus}).

\begin{equation}
\label{eq:floyd}
\textit{path}(a,b) := \textit{path}(a,b) \vee (\textit{path}(a,c) \wedge \textit{path}(c,b))
\end{equation}

After eliminating
$c$, for every path in the original graph which passed through $c$
there is now a shortcut which bypasses $c$. After we have created such
shortcuts for all vertices in the graph it is clear that the resulting
graph has an edge $a \rightarrow b$ iff there is a path from $a$ to
$b$ in the original graph.

\begin{figure}
\begin{center}
\scalebox{.7}{%
\begin{picture}(48,24)
\put(1.25,1){\makebox(0,0)[r]{13}}
\put(1.25,2.5){\makebox(0,0)[r]{12}}
\put(1.25,4){\makebox(0,0)[r]{11}}
\put(1.25,5.5){\makebox(0,0)[r]{10}}
\put(1.25,7){\makebox(0,0)[r]{9}}
\put(1.25,8.5){\makebox(0,0)[r]{8}}
\put(1.25,10){\makebox(0,0)[r]{7}}
\put(1.25,11.5){\makebox(0,0)[r]{6}}
\put(1.25,13){\makebox(0,0)[r]{5}}
\put(1.25,14.5){\makebox(0,0)[r]{4}}
\put(1.25,16){\makebox(0,0)[r]{3}}
\put(1.25,17.5){\makebox(0,0)[r]{2}}
\put(1.25,19){\makebox(0,0)[r]{1}}
\put(2.75,20.75){\makebox(0,0){1}}
\put(4.25,20.75){\makebox(0,0){2}}
\put(5.75,20.75){\makebox(0,0){3}}
\put(7.25,20.75){\makebox(0,0){4}}
\put(8.75,20.75){\makebox(0,0){5}}
\put(10.25,20.75){\makebox(0,0){6}}
\put(11.75,20.75){\makebox(0,0){7}}
\put(13.25,20.75){\makebox(0,0){8}}
\put(14.75,20.75){\makebox(0,0){9}}
\put(16.25,20.75){\makebox(0,0){10}}
\put(17.75,20.75){\makebox(0,0){11}}
\put(19.25,20.75){\makebox(0,0){12}}
\put(20.75,20.75){\makebox(0,0){13}}
\put(15.5,13.75){\textcolor{gray50}{\rectangle{1.5}{1.5}}}
\put(18.5,12.25){\textcolor{gray50}{\rectangle{1.5}{1.5}}}
\put(12.5,9.25){\textcolor{gray50}{\rectangle{1.5}{1.5}}}
\put(3.5,7.75){\textcolor{gray50}{\rectangle{1.5}{1.5}}}
\put(2,6.25){\textcolor{gray50}{\rectangle{1.5}{1.5}}}
\put(2,4.75){\textcolor{gray50}{\rectangle{1.5}{1.5}}}
\put(3.5,3.25){\textcolor{gray50}{\rectangle{1.5}{1.5}}}
\put(5,1.75){\textcolor{gray50}{\rectangle{1.5}{1.5}}}
\put(5,0.25){\textcolor{gray50}{\rectangle{1.5}{1.5}}}
\path(2,0.25)(2,19.75)
\path(3.5,0.25)(3.5,19.75)
\path(5,0.25)(5,19.75)
\path(6.5,0.25)(6.5,19.75)
\path(8,0.25)(8,19.75)
\path(9.5,0.25)(9.5,19.75)
\path(11,0.25)(11,19.75)
\path(12.5,0.25)(12.5,19.75)
\path(14,0.25)(14,19.75)
\path(15.5,0.25)(15.5,19.75)
\path(17,0.25)(17,19.75)
\path(18.5,0.25)(18.5,19.75)
\path(20,0.25)(20,19.75)
\path(21.5,0.25)(21.5,19.75)
\path(2,0.25)(21.5,0.25)
\path(2,1.75)(21.5,1.75)
\path(2,3.25)(21.5,3.25)
\path(2,4.75)(21.5,4.75)
\path(2,6.25)(21.5,6.25)
\path(2,7.75)(21.5,7.75)
\path(2,9.25)(21.5,9.25)
\path(2,10.75)(21.5,10.75)
\path(2,12.25)(21.5,12.25)
\path(2,13.75)(21.5,13.75)
\path(2,15.25)(21.5,15.25)
\path(2,16.75)(21.5,16.75)
\path(2,18.25)(21.5,18.25)
\path(2,19.75)(21.5,19.75)

\put(25.25,1){\makebox(0,0)[r]{13}}
\put(25.25,2.5){\makebox(0,0)[r]{12}}
\put(25.25,4){\makebox(0,0)[r]{11}}
\put(25.25,5.5){\makebox(0,0)[r]{10}}
\put(25.25,7){\makebox(0,0)[r]{9}}
\put(25.25,8.5){\makebox(0,0)[r]{8}}
\put(25.25,10){\makebox(0,0)[r]{7}}
\put(25.25,11.5){\makebox(0,0)[r]{6}}
\put(25.25,13){\makebox(0,0)[r]{5}}
\put(25.25,14.5){\makebox(0,0)[r]{4}}
\put(25.25,16){\makebox(0,0)[r]{3}}
\put(25.25,17.5){\makebox(0,0)[r]{2}}
\put(25.25,19){\makebox(0,0)[r]{1}}
\put(26.75,20.75){\makebox(0,0){1}}
\put(28.25,20.75){\makebox(0,0){2}}
\put(29.75,20.75){\makebox(0,0){3}}
\put(31.25,20.75){\makebox(0,0){4}}
\put(32.75,20.75){\makebox(0,0){5}}
\put(34.25,20.75){\makebox(0,0){6}}
\put(35.75,20.75){\makebox(0,0){7}}
\put(37.25,20.75){\makebox(0,0){8}}
\put(38.75,20.75){\makebox(0,0){9}}
\put(40.25,20.75){\makebox(0,0){10}}
\put(41.75,20.75){\makebox(0,0){11}}
\put(43.25,20.75){\makebox(0,0){12}}
\put(44.75,20.75){\makebox(0,0){13}}
\put(26,13.75){\textcolor{gray50}{\rectangle{1.5}{1.5}}}   
\put(39.5,13.75){\textcolor{gray50}{\rectangle{1.5}{1.5}}} 
\put(29,12.25){\textcolor{gray50}{\rectangle{1.5}{1.5}}}   
\put(42.5,12.25){\textcolor{gray50}{\rectangle{1.5}{1.5}}} 
\put(27.5,9.25){\textcolor{gray50}{\rectangle{1.5}{1.5}}}  
\put(36.5,9.25){\textcolor{gray50}{\rectangle{1.5}{1.5}}}  
\put(27.5,7.75){\textcolor{gray50}{\rectangle{1.5}{1.5}}}  
\put(26,6.25){\textcolor{gray50}{\rectangle{1.5}{1.5}}}    
\put(26,4.75){\textcolor{gray50}{\rectangle{1.5}{1.5}}}    
\put(27.5,3.25){\textcolor{gray50}{\rectangle{1.5}{1.5}}}  
\put(29,1.75){\textcolor{gray50}{\rectangle{1.5}{1.5}}}    
\put(29,0.25){\textcolor{gray50}{\rectangle{1.5}{1.5}}}    
\path(26,0.25)(26,19.75)
\path(27.5,0.25)(27.5,19.75)
\path(29,0.25)(29,19.75)
\path(30.5,0.25)(30.5,19.75)
\path(32,0.25)(32,19.75)
\path(33.5,0.25)(33.5,19.75)
\path(35,0.25)(35,19.75)
\path(36.5,0.25)(36.5,19.75)
\path(38,0.25)(38,19.75)
\path(39.5,0.25)(39.5,19.75)
\path(41,0.25)(41,19.75)
\path(42.5,0.25)(42.5,19.75)
\path(44,0.25)(44,19.75)
\path(45.5,0.25)(45.5,19.75)
\path(26,0.25)(45.5,0.25)
\path(26,1.75)(45.5,1.75)
\path(26,3.25)(45.5,3.25)
\path(26,4.75)(45.5,4.75)
\path(26,6.25)(45.5,6.25)
\path(26,7.75)(45.5,7.75)
\path(26,9.25)(45.5,9.25)
\path(26,10.75)(45.5,10.75)
\path(26,12.25)(45.5,12.25)
\path(26,13.75)(45.5,13.75)
\path(26,15.25)(45.5,15.25)
\path(26,16.75)(45.5,16.75)
\path(26,18.25)(45.5,18.25)
\path(26,19.75)(45.5,19.75)
\Thicklines
\path(26,15.25)(30.5,15.25)(30.5,10.75)(26,10.75)(26,15.25)
\end{picture}}
\end{center}

\caption{Initial graph (left) and its transitive closure (right)}
\label{fig:transc}
\end{figure}
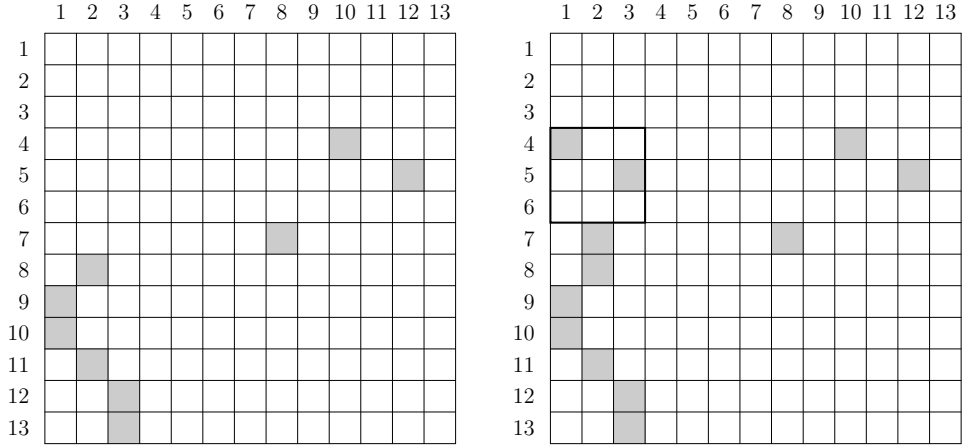

Figure~\vref{fig:transc} shows the essential net of
Figure~\vref{fig:ensomeoneleft} in adjacency matrix representation
(left of the figure) and its transitive closure (right of the figure).
A square in the matrix is colored in iff there is a link from the row
to the column in the graph.

The relevant part of the graph for the acyclicity test is marked by a square around
columns $1-3$ of row $4-6$. We see here, for example, that given the
existence of a path $4\rightarrow 1$ an axiom link between $s_1$ and
$s_4$ would produce a cycle (via node 10 in the original graph) and is
therefore to be excluded. A similar observation can be made for $s_5$
and $s_3$.

\editout{
A final improvement is possible by taking into avoiding recomputation
of the transitive closure of each link and taking into account the
fact that only a single edge has been added to the
graph. \citet{fullydyn} propose a relatively simple algorithm for
maintaining the transitive closure of a graph under additions and
deletions of edges by keeping track of the \emph{number} of paths
between two points and using additions/subtractions for the
updates. Their algorithm has constant query time, $O(n^3)$ initialization time
and can update the transitive closure in $O(n^2)$ if all added/deleted
edges are incident on a single vertex. Though this algorithm depends
on constant time arithmetic 

For the current application, however, simple inspection of the links
for essential nets shows us that only a negative product link
introduces a vertex with two paths leaving from it, while all the
other links only have vertices with at most on edge leaving from it.
This means there can be at most $p+1$ paths between two points, where
$p$ is the number of negative product links. Given that this number is
unlikely to be 
}
\section{Connectedness}

Verifying conditions (2) and (3) from Definition \ref{def:correctdyn}
is a bit harder. The question we want to ask about each link is: does
this link contribute to a connected proof structure? Or, inversely, does
excluding the other possibilities for the two atomic formulas we
connect mean a connected proof structure is still possible.

To check the conditions we need to verify the following:

\begin{enumerate}
\item for every negative input of the net we verify there exists
  a path to the positive conclusion,
\item for every negative $\bullet$ link we verify that \emph{both}
  paths leaving from it reach their destination,
\item for every positive $/$ or $\bs$ link we check the existence of a
  path from its negative premiss to its positive conclusion continuing
  to the positive conclusion of the essential net.
\end{enumerate}

Given that we are already computing the transitive closure of the
graph for verifying acyclicity, we can exploit this by adding
additional information to the matrix we use for the transitive
closure. There are many ways of storing this extra information, the
simplest being in the form of an ordered list of pairs. Given that, for
$a$ atomic formulas, each possible connection allows $(a-1)^2$ other
connections (ie.\ it is agnostic about all possibilities not
contradicting this one) but excludes $2(a-1)$ possibilities, it is
more economic to store the connections which are excluded.
For example, the ordered set associated to the edge from 1 to 4 will
be $\{1-5, 1-6, 2-4, 3-4\}$, meaning ``there is an edge from 1 to 4
but not to anywhere else and the only edge arriving at 4 comes from 1''.

Note that in the description of the Floyd-Warshall algorithm, we made
use only of the logical `and' and `or' operators. For ordered
sets, the corresponding operations are set union and set intersection.
For eliminating vertex $c$ from a path from $a$ to $b$, we first take
the union of the ordered set representing the links which are
not in a path from $a$ to $c$ with that representing the links not in
a path from $c$ to $b$ (any vertex in either set couldn't be in a path
from $a$ via $c$ to $b$). Then, we take the intersection of this set
with the old set associated to the path from $a$ to $b$.

\begin{equation}
\label{eq:floydinv}
\overline{\textit{path}}(a,b) := \overline{\textit{path}}(a,b) \cap (\overline{\textit{path}}(a,c) \cup \overline{\textit{path}}(c,b))
\end{equation}

Note that Equation~\ref{eq:floydinv} is simply Equation~\ref{eq:floyd}
with both sides negated, the negations moved inward and set union and
intersection in the place of the logical `or' and `and' operators.

Given that we can implement the union and intersection operations in
linear time with respect to the size of the input sets, the total
complexity of our algorithm becomes $O(v^3 2(a-1)) = O(a^4)$.

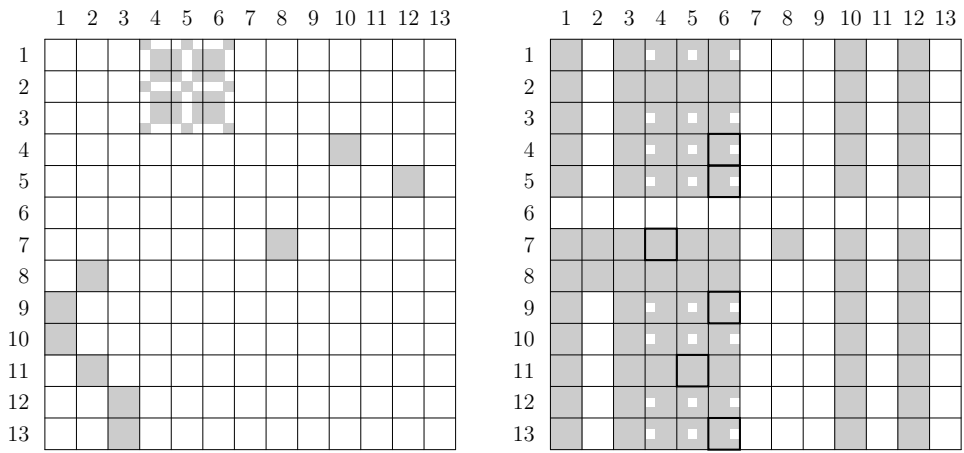
\begin{figure}
\begin{center}
\scalebox{.7}{%
\begin{picture}(48,24)
\put(1.25,1){\makebox(0,0)[r]{13}}
\put(1.25,2.5){\makebox(0,0)[r]{12}}
\put(1.25,4){\makebox(0,0)[r]{11}}
\put(1.25,5.5){\makebox(0,0)[r]{10}}
\put(1.25,7){\makebox(0,0)[r]{9}}
\put(1.25,8.5){\makebox(0,0)[r]{8}}
\put(1.25,10){\makebox(0,0)[r]{7}}
\put(1.25,11.5){\makebox(0,0)[r]{6}}
\put(1.25,13){\makebox(0,0)[r]{5}}
\put(1.25,14.5){\makebox(0,0)[r]{4}}
\put(1.25,16){\makebox(0,0)[r]{3}}
\put(1.25,17.5){\makebox(0,0)[r]{2}}
\put(1.25,19){\makebox(0,0)[r]{1}}
\put(2.75,20.75){\makebox(0,0){1}}
\put(4.25,20.75){\makebox(0,0){2}}
\put(5.75,20.75){\makebox(0,0){3}}
\put(7.25,20.75){\makebox(0,0){4}}
\put(8.75,20.75){\makebox(0,0){5}}
\put(10.25,20.75){\makebox(0,0){6}}
\put(11.75,20.75){\makebox(0,0){7}}
\put(13.25,20.75){\makebox(0,0){8}}
\put(14.75,20.75){\makebox(0,0){9}}
\put(16.25,20.75){\makebox(0,0){10}}
\put(17.75,20.75){\makebox(0,0){11}}
\put(19.25,20.75){\makebox(0,0){12}}
\put(20.75,20.75){\makebox(0,0){13}}
\put(6.5,18.25){\ynnnyynyy{gray50}}
\put(8,18.25){\nynynyyny{gray50}}
\put(9.5,18.25){\nnyyynyyn{gray50}}
\put(6.5,16.75){\nyyynnnyy{gray50}}
\put(8,16.75){\ynynynyny{gray50}}
\put(9.5,16.75){\yynnnyyyn{gray50}}
\put(6.5,15.25){\nyynyyynn{gray50}}
\put(8,15.25){\ynyynynyn{gray50}}
\put(9.5,15.25){\yynyynnny{gray50}}
\put(15.5,13.75){\textcolor{gray50}{\rectangle{1.5}{1.5}}}
\put(18.5,12.25){\textcolor{gray50}{\rectangle{1.5}{1.5}}}
\put(12.5,9.25){\textcolor{gray50}{\rectangle{1.5}{1.5}}}
\put(3.5,7.75){\textcolor{gray50}{\rectangle{1.5}{1.5}}}
\put(2,6.25){\textcolor{gray50}{\rectangle{1.5}{1.5}}}
\put(2,4.75){\textcolor{gray50}{\rectangle{1.5}{1.5}}}
\put(3.5,3.25){\textcolor{gray50}{\rectangle{1.5}{1.5}}}
\put(5,1.75){\textcolor{gray50}{\rectangle{1.5}{1.5}}}
\put(5,0.25){\textcolor{gray50}{\rectangle{1.5}{1.5}}}
\path(2,0.25)(2,19.75)
\path(3.5,0.25)(3.5,19.75)
\path(5,0.25)(5,19.75)
\path(6.5,0.25)(6.5,19.75)
\path(8,0.25)(8,19.75)
\path(9.5,0.25)(9.5,19.75)
\path(11,0.25)(11,19.75)
\path(12.5,0.25)(12.5,19.75)
\path(14,0.25)(14,19.75)
\path(15.5,0.25)(15.5,19.75)
\path(17,0.25)(17,19.75)
\path(18.5,0.25)(18.5,19.75)
\path(20,0.25)(20,19.75)
\path(21.5,0.25)(21.5,19.75)
\path(2,0.25)(21.5,0.25)
\path(2,1.75)(21.5,1.75)
\path(2,3.25)(21.5,3.25)
\path(2,4.75)(21.5,4.75)
\path(2,6.25)(21.5,6.25)
\path(2,7.75)(21.5,7.75)
\path(2,9.25)(21.5,9.25)
\path(2,10.75)(21.5,10.75)
\path(2,12.25)(21.5,12.25)
\path(2,13.75)(21.5,13.75)
\path(2,15.25)(21.5,15.25)
\path(2,16.75)(21.5,16.75)
\path(2,18.25)(21.5,18.25)
\path(2,19.75)(21.5,19.75)

\put(25.25,1){\makebox(0,0)[r]{13}}
\put(25.25,2.5){\makebox(0,0)[r]{12}}
\put(25.25,4){\makebox(0,0)[r]{11}}
\put(25.25,5.5){\makebox(0,0)[r]{10}}
\put(25.25,7){\makebox(0,0)[r]{9}}
\put(25.25,8.5){\makebox(0,0)[r]{8}}
\put(25.25,10){\makebox(0,0)[r]{7}}
\put(25.25,11.5){\makebox(0,0)[r]{6}}
\put(25.25,13){\makebox(0,0)[r]{5}}
\put(25.25,14.5){\makebox(0,0)[r]{4}}
\put(25.25,16){\makebox(0,0)[r]{3}}
\put(25.25,17.5){\makebox(0,0)[r]{2}}
\put(25.25,19){\makebox(0,0)[r]{1}}
\put(26.75,20.75){\makebox(0,0){1}}
\put(28.25,20.75){\makebox(0,0){2}}
\put(29.75,20.75){\makebox(0,0){3}}
\put(31.25,20.75){\makebox(0,0){4}}
\put(32.75,20.75){\makebox(0,0){5}}
\put(34.25,20.75){\makebox(0,0){6}}
\put(35.75,20.75){\makebox(0,0){7}}
\put(37.25,20.75){\makebox(0,0){8}}
\put(38.75,20.75){\makebox(0,0){9}}
\put(40.25,20.75){\makebox(0,0){10}}
\put(41.75,20.75){\makebox(0,0){11}}
\put(43.25,20.75){\makebox(0,0){12}}
\put(44.75,20.75){\makebox(0,0){13}}
\put(26,18.25){\yyyyyyyyy{gray50}}     
\put(29,18.25){\yyyyyyyyy{gray50}}     
\put(30.5,18.25){\yyynyyyyy{gray50}}   
\put(32,18.25){\yyyynyyyy{gray50}}     
\put(33.5,18.25){\yyyyynyyy{gray50}}   
\put(39.5,18.25){\yyyyyyyyy{gray50}}   
\put(42.5,18.25){\yyyyyyyyy{gray50}}   
\put(26,16.75){\yyyyyyyyy{gray50}}     
\put(29,16.75){\yyyyyyyyy{gray50}}     
\put(30.5,16.75){\yyyyyyyyy{gray50}}   
\put(32,16.75){\yyyyyyyyy{gray50}}     
\put(33.5,16.75){\yyyyyyyyy{gray50}}   
\put(39.5,16.75){\yyyyyyyyy{gray50}}   
\put(42.5,16.75){\yyyyyyyyy{gray50}}   
\put(26,15.25){\yyyyyyyyy{gray50}}     
\put(29,15.25){\yyyyyyyyy{gray50}}     
\put(30.5,15.25){\yyynyyyyy{gray50}}   
\put(32,15.25){\yyyynyyyy{gray50}}     
\put(33.5,15.25){\yyyyynyyy{gray50}}   
\put(39.5,15.25){\yyyyyyyyy{gray50}}   
\put(42.5,15.25){\yyyyyyyyy{gray50}}   
\put(26,13.75){\yyyyyyyyy{gray50}}     
\put(29,13.75){\yyyyyyyyy{gray50}}     
\put(30.5,13.75){\yyynyyyyy{gray50}}   
\put(32,13.75){\yyyynyyyy{gray50}}     
\put(33.5,13.75){\yyyyynyyy{gray50}}   
\put(39.5,13.75){\yyyyyyyyy{gray50}}   
\put(42.5,13.75){\yyyyyyyyy{gray50}}   
\put(26,12.25){\yyyyyyyyy{gray50}}     
\put(29,12.25){\yyyyyyyyy{gray50}}     
\put(30.5,12.25){\yyynyyyyy{gray50}}   
\put(32,12.25){\yyyynyyyy{gray50}}     
\put(33.5,12.25){\yyyyynyyy{gray50}}   
\put(39.5,12.25){\yyyyyyyyy{gray50}}   
\put(42.5,12.25){\yyyyyyyyy{gray50}}   
\put(26,9.25){\yyyyyyyyy{gray50}}      
\put(27.5,9.25){\yyyyyyyyy{gray50}}    
\put(29,9.25){\yyyyyyyyy{gray50}}      
\put(30.5,9.25){\yyyyyyyyy{gray50}}    
\put(32,9.25){\yyyyyyyyy{gray50}}      
\put(33.5,9.25){\yyyyyyyyy{gray50}}    
\put(36.5,9.25){\yyyyyyyyy{gray50}}    
\put(39.5,9.25){\yyyyyyyyy{gray50}}    
\put(42.5,9.25){\yyyyyyyyy{gray50}}    
\put(26,7.75){\yyyyyyyyy{gray50}}      
\put(27.5,7.75){\yyyyyyyyy{gray50}}    
\put(29,7.75){\yyyyyyyyy{gray50}}      
\put(30.5,7.75){\yyyyyyyyy{gray50}}    
\put(32,7.75){\yyyyyyyyy{gray50}}      
\put(33.5,7.75){\yyyyyyyyy{gray50}}    
\put(39.5,7.75){\yyyyyyyyy{gray50}}    
\put(42.5,7.75){\yyyyyyyyy{gray50}}    
\put(26,6.25){\yyyyyyyyy{gray50}}      
\put(29,6.25){\yyyyyyyyy{gray50}}      
\put(30.5,6.25){\yyynyyyyy{gray50}}    
\put(32,6.25){\yyyynyyyy{gray50}}      
\put(33.5,6.25){\yyyyynyyy{gray50}}    
\put(39.5,6.25){\yyyyyyyyy{gray50}}    
\put(42.5,6.25){\yyyyyyyyy{gray50}}    
\put(26,4.75){\yyyyyyyyy{gray50}}      
\put(29,4.75){\yyyyyyyyy{gray50}}      
\put(30.5,4.75){\yyynyyyyy{gray50}}    
\put(32,4.75){\yyyynyyyy{gray50}}      
\put(33.5,4.75){\yyyyynyyy{gray50}}    
\put(39.5,4.75){\yyyyyyyyy{gray50}}    
\put(42.5,4.75){\yyyyyyyyy{gray50}}    
\put(26,3.25){\yyyyyyyyy{gray50}}     
\put(29,3.25){\yyyyyyyyy{gray50}}     
\put(30.5,3.25){\yyyyyyyyy{gray50}}   
\put(32,3.25){\yyyyyyyyy{gray50}}     
\put(33.5,3.25){\yyyyyyyyy{gray50}}   
\put(39.5,3.25){\yyyyyyyyy{gray50}}   
\put(42.5,3.25){\yyyyyyyyy{gray50}}   
\put(26,1.75){\yyyyyyyyy{gray50}}      
\put(29,1.75){\yyyyyyyyy{gray50}}      
\put(30.5,1.75){\yyynyyyyy{gray50}}    
\put(32,1.75){\yyyynyyyy{gray50}}      
\put(33.5,1.75){\yyyyynyyy{gray50}}    
\put(39.5,1.75){\yyyyyyyyy{gray50}}    
\put(42.5,1.75){\yyyyyyyyy{gray50}}    
\put(26,0.25){\yyyyyyyyy{gray50}}      
\put(29,0.25){\yyyyyyyyy{gray50}}      
\put(30.5,0.25){\yyynyyyyy{gray50}}    
\put(32,0.25){\yyyynyyyy{gray50}}      
\put(33.5,0.25){\yyyyynyyy{gray50}}    
\put(39.5,0.25){\yyyyyyyyy{gray50}}    
\put(42.5,0.25){\yyyyyyyyy{gray50}}    
\path(26,0.25)(26,19.75)
\path(27.5,0.25)(27.5,19.75)
\path(29,0.25)(29,19.75)
\path(30.5,0.25)(30.5,19.75)
\path(32,0.25)(32,19.75)
\path(33.5,0.25)(33.5,19.75)
\path(35,0.25)(35,19.75)
\path(36.5,0.25)(36.5,19.75)
\path(38,0.25)(38,19.75)
\path(39.5,0.25)(39.5,19.75)
\path(41,0.25)(41,19.75)
\path(42.5,0.25)(42.5,19.75)
\path(44,0.25)(44,19.75)
\path(45.5,0.25)(45.5,19.75)
\path(26,0.25)(45.5,0.25)
\path(26,1.75)(45.5,1.75)
\path(26,3.25)(45.5,3.25)
\path(26,4.75)(45.5,4.75)
\path(26,6.25)(45.5,6.25)
\path(26,7.75)(45.5,7.75)
\path(26,9.25)(45.5,9.25)
\path(26,10.75)(45.5,10.75)
\path(26,12.25)(45.5,12.25)
\path(26,13.75)(45.5,13.75)
\path(26,15.25)(45.5,15.25)
\path(26,16.75)(45.5,16.75)
\path(26,18.25)(45.5,18.25)
\path(26,19.75)(45.5,19.75)
\Thicklines
\path(33.5,15.25)(35,15.25)(35,13.75)(33.5,13.75)(33.5,15.25) 
\path(33.5,13.75)(35,13.75)(35,12.25)(33.5,12.25)(33.5,13.75) 
\path(30.5,10.75)(32,10.75)(32,9.25)(30.5,9.25)(30.5,10.75)   
\path(33.5,7.75)(35,7.75)(35,6.25)(33.5,6.25)(33.5,7.75)      
\path(32,4.75)(33.5,4.75)(33.5,3.25)(32,3.25)(32,4.75)      
\path(33.5,1.75)(35,1.75)(35,0.25)(33.5,0.25)(33.5,1.75)      
\end{picture}}
\end{center}

\caption{Initial graph (left) and its transitive closure (right)}
\label{fig:transcb}
\end{figure}

Figure~\vref{fig:transcb} shows the initial graph and its transitive
closure. Every entry from the previous graph is now divided into 9
subentries --- one for each possible axiom link. We read the entry
for $1-4$ as follows: the first row indicate the link between 1 and 4,
and thereby the absence of a link $1-5$ and $1-6$, the second row
indicates the possibilities for linking 2, which just excludes $2-4$
and the third row indicates that for 3 just the $3-4$ connection is
impossible. Again, we have marked
the table entries
relevant for our correctness condition by drawing a black border
around them.

When we look at the transitive closure, we see that,
should we choose to link $s_2$ to $s_6$, this would make it impossible
to reach vertex 6 from vertices 4, 5, 9 or 13. Remark also that cycles need to be excluded separately. For example, the path from 5 to 1 in
Figure~\ref{fig:transcb} does not mean we need to exclude the $s_1-s_5$
axiom link.

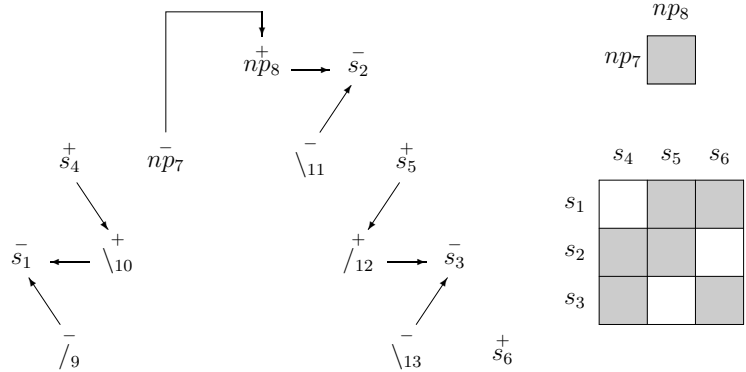
\begin{figure}
 \begin{center}
  \scalebox{0.8}{
  \begin{picture}(34,16)
   \put(2,1){\trldra{\ant{\mathbin{/_9}}}{\ant{s_1}}{\suc{\mathbin{\bs_{10}}}}}
   \put(4,5){\trrdr{}{\suc{s_4}}{\ant{np_7}}}

   \put(12,9){\trldla{\ant{\mathbin{\bs_{11}}}}{\suc{np_8}}{\ant{s_2}}}
   \put(14,5){\trrdl{\suc{\mathbin{/_{12}}}}{}{\suc{s_5}}}
   \put(16,1){\trldla{\ant{\mathbin{\bs_{13}}}}{}{\ant{s_3}}}

   \put(22,1){\makebox(0,0){\suc{s_6}}}
   \path(8,10)(8,15)(12,15)
   \put(12,15){\vector(0,-1){1}}

\put(28,6){\textcolor{gray50}{\rectangle{4}{2}}}
\put(26,4){\textcolor{gray50}{\rectangle{4}{2}}}
\put(26,2){\textcolor{gray50}{\rectangle{2}{2}}}
\put(30,2){\textcolor{gray50}{\rectangle{2}{2}}}
\path(26,2)(32,2)
\path(26,4)(32,4)
\path(26,6)(32,6)
\path(26,8)(32,8)
\path(26,2)(26,8)
\path(28,2)(28,8)
\path(30,2)(30,8)
\path(32,2)(32,8)
\put(25,7){\makebox(0,0){$s_1$}}
\put(25,5){\makebox(0,0){$s_2$}}
\put(25,3){\makebox(0,0){$s_3$}}
\put(27,9){\makebox(0,0){$s_4$}}
\put(29,9){\makebox(0,0){$s_5$}}
\put(31,9){\makebox(0,0){$s_6$}}

\put(28,12){\textcolor{gray50}{\rectangle{2}{2}}}
\put(29,15){\makebox(0,0){$np_8$}}
\put(27,13){\makebox(0,0){$np_7$}}
\path(28,12)(28,14)(30,14)(30,12)(28,12)
  \end{picture}}
 \end{center}

\caption{Essential net with acyclicity and connectedness taken into account}
\label{fig:ensomeoneleftb}
\end{figure}

Figure~\ref{fig:ensomeoneleftb} shows the proof frame of
Figure~\ref{fig:ensomeoneleft} with all constraints taken into account. We
see that, whatever choice we make for the first axiom link, all
other axiom links will be fixed immediately, giving us the two proofs
$s_1-s_5$, $s_2-s_4$, $s_3-s_6$ (ie.\ the proof shown in
Figure~\ref{fig:pnsomeoneleft}) and $s_1-s_6$, $s_2-s_5$, $s_3-s_4$.

\section{Extensions and Improvements}

An interesting continuation of the themes explored in this paper would be to
look at \emph{dynamic} graph algorithms, where we maintain the
transitive closure under additions and deletions of
edges. This would avoid recomputing the transitive closure from
scratch after every axiom link and would allow us to take advantage of the
information we have already computed. \citet{fullydyn} propose an
algorithm with $O(n^2)$ update time based on keeping track of the
number of paths between two vertices, which is easy enough to adapt to
our current scenario in the case of acyclicity tests, though it
remains unclear if it can be adopted to check for connectedness.

Another improvement would be to represent the ordered sets
differently. Given that their structure is quite regular, it may be 
possible to improve upon linear time union and intersection. However,
given that for each iteration the size of the
sets either remains the same or decreases (the principal operation in
Equation~\ref{eq:floydinv} being intersection), it remains to be seen
if this will result in a practical improvement.

Finally, we can consider the work in the two previous sections as using a sort of `lookahead' of one axiom link, which is to say we exclude all axioms
links which, by themselves, would produce a cyclic or disconnected
proof structure. This can be extended quite naturally to doing $k$
axiom links at the same time, though each extra axiom link will
multiply the required space by $O(n^2)$, the required time for
acyclicity by $O(n^3)$ and the required time for connectedness by
$O(n^4)$ (this is relatively easy to see because we are in effect
substituting $n (n-1)$ for the old value of $n$).

\section{Polynomial Time}

If we add weights to the different connections, the situation
changes. The simplest way to add weights to our graph is to use the
\emph{distance} between two atomic formulas as their weight and
prefer the total axiom linking with the least total weight. This
choice of assigning weights is closely related to work on left-to-right
processing of sentences, proposed independently by \cite{pnproc} and
\citet{incrpro}. The
claim they make is that the complexity of a phrase depends on the
number of `open' or unlinked axiom formulas a reader/listener will
have to maintain in memory to produce a parse for this sentence.

Finding a minimum-weight solution to this problem is known as the
\emph{assignment problem}. \citet{murty68assign} was the first to give
an algorithm for generating the assignments in order of increasing cost. His
$O(kn^4)$ algorithm for finding the $k$ best assignments can be
improved to $O(kn^3)$, even though tests on randomly generated graphs
have shown the observed complexity to be $O(kn^2)$  \citep{miller97optim}.

\begin{figure}
 \begin{center}
 \scalebox{0.8}{
  \begin{picture}(38,18)
   \put(0,2){\makebox(0,0){\suc{s_1}}}

   \put(2,2){\trldra{\ant{/}}{\ant{s_2}}}
   \put(4,6){\trrdr{\suc{\bs}}{\suc{s_3}}{\ant{np_7}}}

   \put(10,6){\trldla{\ant{\bs}}{\suc{np_8}}{\ant{s_4}}}
   \put(12,2){\trldra{\ant{/}}{}{\suc{np_9}}}

   \put(18,6){\trrdl{\suc{\bs}}{\ant{np_{10}}}{\ant{s_5}}}
   \put(20,2){\trldla{\ant{/}}{}{\ant{s_6}}}
   
   \put(4,0){\makebox(0,0){\textit{someone}}}
   \put(14,0){\makebox(0,0){\textit{loves}}}
   \put(22,0){\makebox(0,0){\textit{everyone}}}

   \drawline(8,11)(8,12)(10,12)          
   \drawline(16,12)(18,12)(18,11)        
   \drawline(0,8)(2,8)(2,7)              
   \drawline(22,13)(14,13)(14,11)        
   \drawline(4,11)(4,14)(24,14)          
   \put(10,12){\vector(0,-1){1}}
   \put(16,12){\vector(0,-1){5}}
   \put(0,8){\vector(0,-1){5}}
   \put(22,13){\vector(0,-1){2}}
   \put(24,14){\vector(0,-1){7}}

\put(32,8){\textcolor{gray80}{\rectangle{2}{2}}}
\put(34,10){\textcolor{gray80}{\rectangle{2}{2}}}
\put(30,12){\textcolor{gray80}{\rectangle{2}{2}}}
\path(30,8)(36,8)
\path(30,10)(36,10)
\path(30,12)(36,12)
\path(30,14)(36,14)
\path(30,8)(30,14)
\path(32,8)(32,14)
\path(34,8)(34,14)
\path(36,8)(36,14)

\put(31,2){\textcolor{gray80}{\rectangle{2}{2}}}
\put(33,0){\textcolor{gray80}{\rectangle{2}{2}}}
\put(29.8,3){\cmput{np_7}}
\put(29.8,1){\cmput{np_9}}
\put(32,5){\cmput{np_8}}
\put(34,5){\cmput{np_{10}}}
\put(32,1){\cmput{3}}
\put(34,1){\cmput{1}}
\put(32,3){\cmput{1}}
\put(34,3){\cmput{3}}

\path(31,0)(31,4)
\path(33,0)(33,4)
\path(35,0)(35,4)
\path(31,0)(35,0)
\path(31,2)(35,2)
\path(31,4)(35,4)

\put(29,9){\cmput{s_6}}
\put(29,11){\cmput{s_4}}
\put(29,13){\cmput{s_2}}

\put(31,15){\cmput{s_1}}
\put(33,15){\cmput{s_3}}
\put(35,15){\cmput{s_5}}

\put(31,13){\cmput{1}}
\put(31,11){\cmput{\infty}}
\put(31,9){\cmput{9}}

\put(33,13){\cmput{\infty}}
\put(33,11){\cmput{3}}
\put(33,9){\cmput{7}}

\put(35,13){\cmput{7}}
\put(35,11){\cmput{3}}
\put(35,9){\cmput{\infty}}

  \end{picture}}
 \end{center}
\caption{Minimum Weight Linking for `someone loves everyone'}
\label{fig:leftright}
\end{figure}
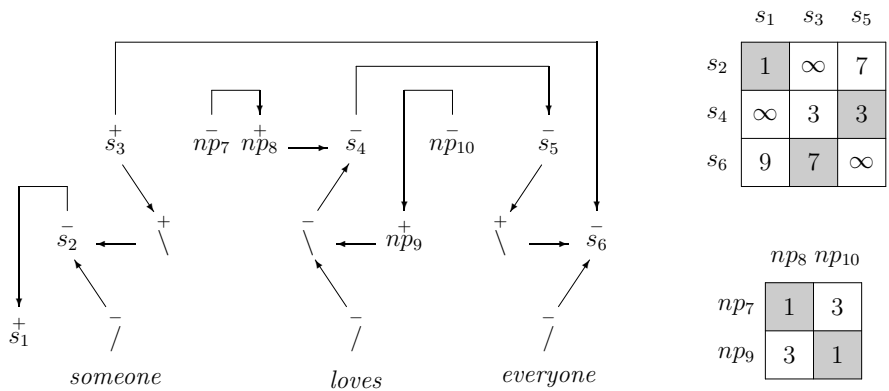

Because using the distance as weight tends to favor cyclic
connections, it is preferable to make one pass of the algorithm
described in the previous section and assign a weight of infinity to
all edges which are either cyclic or
disconnected. Figure~\ref{fig:leftright} shows a weighted graph
corresponding to an example from \cite{incrpro}, the sentence `someone
loves everyone', which has a preferred reading where the subject has
wide scope. Of the four readings of this sentence (two if we enforce
planarity) there is a preference for connecting $s_1-s_2$, $s_3-s_6$,
$s_5-s_4$, with a total weight of 11, as compared to connecting
$s_1-s_6$, $s_3-s_4$, $s_5-s_2$, with a total weight of 19.

There is an important difference between using distance weights
like we do here and keeping track of the open axiom links like Johnson
and Morrill, which is that we select a linking which is best
\emph{globally}. It is therefore to be expected that we will make different
predictions for some `garden path'
sentences (ie.\ sentences where a suboptimal local choice for the
axiom links will be made).

Finding an appropriate value of $k$ and fragments of type-logical
grammar for which this $k$ is guaranteed to find all readings remains
an interesting open question. Cautious people might select $k = n!$,
given that in any type-logical grammar there are at most $n!$ links
possible, to generate all readings in increasing order of complexity. It seems
tempting to set $k$ to $n^3$, because many interesting grammar
formalisms have $O(n^6)$ complexity \citep{conv}, and find type-logical fragments
for which we can show proof search using this strategy is complete.

\section{Conclusion}

We have seen how standard graph algorithms can be modified to aid in
proof search for type-logical grammars by rejecting connections which
can never contribute to a successful proof.

We have also seen how weighing the connections allows us to enumerate
the links in increasing order and linked this with processing claims.




\bibliography{moot}
\bibliographystyle{elsart-harv}

\end{document}

%% file: moot_pnmacros.tex
 \newcounter{myovalx}
 \newcounter{myovaly}
 
 \newcounter{myovalxx}
 \newcounter{myovalyy}

 \newcounter{dotfactor}
\newcommand{\editout}[1]{}

\newcommand{\weg}[1]{}

\newcommand{\rectangle}[2]{%
 \rule{#1\unitlength}{#2\unitlength}}

\newcommand{\cmput}[1]{%
 \makebox(0,0){\ensuremath{#1}}}

\newcommand{\bo}{[}
\newcommand{\bc}{]}
\newcommand{\bs}{\backslash}
\newcommand{\Gammap}{\Gamma^{\prime}}

\newcommand{\ra}{\vdash\rule{0mm}{0.6em}}

\newcommand{\timeslink}[3]{%
\mbox{%
  \begin{picture}(4,4)
  \put(0,4){\makebox(0,0){$ #2 $} }  
  \put(4,4){\makebox(0,0){$ #3 $} }  
  \put(2,0){\makebox(0,0){$ #1 $} }
  \drawline(2.4,1)(3.9,3)
  \drawline(1.6,1)(0.1,3)
  \end{picture}}}
\newcommand{\leftlink}[3]{%
\mbox{%
  \begin{picture}(4,4)
  \put(0,4){\makebox(0,0){$ #2 $} }
  \put(4,4){\makebox(0,0){$ #3 $} }
  \put(2,0){\makebox(0,0){$ #1 $} }
  \drawline(1.6,1)(0.1,3)
  \end{picture}}}
\newcommand{\rightlink}[3]{%
\mbox{%
  \begin{picture}(4,4)
  \put(0,4){\makebox(0,0){$ #2 $} }
  \put(4,4){\makebox(0,0){$ #3 $} }
  \put(2,0){\makebox(0,0){$ #1 $} }
  \drawline(2.4,1)(3.9,3)
  \end{picture}}}

\newcommand{\parlink}[3]{%
\mbox{%
  \begin{picture}(4,4)
  \put(0,4){\makebox(0,0){$ #2 $} }  
  \put(4,4){\makebox(0,0){$ #3 $} }  
  \put(2,0){\makebox(0,0){$ #1 $} }
  \dottedline{.2}(2.4,1)(3.9,3)
  \dottedline{.2}(1.6,1)(0.1,3)
  \qbezier(2.4,1)(2,1.5)(1.6,1)
  \end{picture}}}

\newcommand{\trlprod}[3]{%
\mbox{%
  \begin{picture}(4,4)
  \put(0,4){\makebox(0,0){$ #2 $} }  
  \put(4,4){\makebox(0,0){$ #3 $} }  
  \put(2,0){\makebox(0,0){$ #1 $} }
  \put(2.4,1){\vector(2,3){1.3}}
  \put(1.6,1){\vector(-2,3){1.3}}
  \end{picture}}}

\newcommand{\trrprod}[3]{%
\mbox{%
  \begin{picture}(4,4)
  \put(0,4){\makebox(0,0){$ #2 $} }  
  \put(4,4){\makebox(0,0){$ #3 $} }  
  \put(2,0){\makebox(0,0){$ #1 $} }
  \put(3.7,2.9){\vector(-2,-3){1.3}}
  \put(0.3,2.9){\vector(2,-3){1.3}}
  \end{picture}}}

\newcommand{\trldra}[3]{%
\mbox{%
  \begin{picture}(4,4)
  \put(0,4){\makebox(0,0){$ #2 $} }  
  \put(4,4){\makebox(0,0){$ #3 $} }  
  \put(2,0){\makebox(0,0){$ #1 $} }
  \put(1.6,1){\vector(-2,3){1.3}}
  \put(2.8,3.6){\vector(-1,0){1.6}}
  \end{picture}}}

\newcommand{\trldla}[3]{%
\mbox{%
  \begin{picture}(4,4)
  \put(0,4){\makebox(0,0){$ #2 $} }  
  \put(4,4){\makebox(0,0){$ #3 $} }  
  \put(2,0){\makebox(0,0){$ #1 $} }
  \put(2.4,1){\vector(2,3){1.3}}
  \put(1.2,3.6){\vector(1,0){1.6}}
  \end{picture}}}

\newcommand{\trrdr}[3]{%
\mbox{%
  \begin{picture}(4,4)
  \put(0,4){\makebox(0,0){$ #2 $} }  
  \put(4,4){\makebox(0,0){$ #3 $} }  
  \put(2,0){\makebox(0,0){$ #1 $} }
  \put(0.3,2.9){\vector(2,-3){1.3}}
  \end{picture}}}

\newcommand{\trrdl}[3]{%
\mbox{%
  \begin{picture}(4,4)
  \put(0,4){\makebox(0,0){$ #2 $} }  
  \put(4,4){\makebox(0,0){$ #3 $} }  
  \put(2,0){\makebox(0,0){$ #1 $} }
  \put(3.7,2.9){\vector(-2,-3){1.3}}
  \end{picture}}}

\newcommand{\ant}[1]{\ensuremath{\stackrel{-}{#1}}}
\newcommand{\suc}[1]{\ensuremath{\stackrel{+}{#1}}}

\newcommand{\delete}[1]{}

\newcommand{\cals}{\mathcal{S}}